\documentclass[10pt,twocolumn,letterpaper]{article}

\usepackage{cvpr}
\usepackage{times}
\usepackage{epsfig}
\usepackage{graphicx}
\usepackage{amsmath}
\usepackage{amssymb}
\usepackage{threeparttable}
\usepackage{multirow}
\usepackage{booktabs}
\usepackage{array}
\usepackage{subcaption}
\graphicspath{{pics/}}

\usepackage[breaklinks=true,bookmarks=false]{hyperref}

\cvprfinalcopy 

\def\cvprPaperID{5116} 

\ifcvprfinal\fi
\begin{document}

\title{LT-Net: Label Transfer by Learning Reversible Voxel-wise Correspondence for One-shot Medical Image Segmentation}

\author{
Shuxin~Wang\footnotemark[1]\ \ $^{1,2}$, Shilei~Cao\footnotemark[1]\ \ $^{2}$, Dong~Wei\footnotemark[1]\ \ $^{2}$, Renzhen~Wang$^{4}$, Kai Ma$^{2}$, \\
Liansheng Wang\footnotemark[2]\ \ $^{1}$, Deyu Meng$^{3,4}$ and Yefeng Zheng\footnotemark[2]\ \ $^{2}$\\
\\
$^{1}$ Xiamen University $^{2}$ Jarvis Lab, Tencent \\
 $^{3}$ Macau University of Science and Technology $^{4}$ Xi'an Jiaotong University \\
{\small{sxwang@stu.xmu.edu.cn, lswang@xmu.edu.cn}, \small\{eliasslcao, donwei, kylekma, yefengzheng\}@tencent.com}, \\ \small{wrzhen@stu.xjtu.edu.cn,dymeng@mail.xjtu.edu.cn}
}

\maketitle
\thispagestyle{empty}
\renewcommand{\thefootnote}{\fnsymbol{footnote}}
\footnotetext[1]{Equal contributions. Shuxin Wang contributed to this work during an internship at Tencent.}
\footnotetext[2]{Corresponding authors.}
\renewcommand{\thefootnote}{\arabic{footnote}}

\begin{abstract}
We introduce a one-shot segmentation method to alleviate the burden of manual annotation for medical images.
The main idea is to treat one-shot segmentation as a classical atlas-based segmentation problem,\footnote{We follow Zhao \emph{et al.}~\cite{zhao2019dataaug} to use the phrase ``one-shot'' for its generalized meaning that only one exemplar annotation is needed for the proposed model to learn to segment medical images.} where voxel-wise correspondence from the atlas to the unlabelled data is learned.
Subsequently, segmentation label of the atlas can be transferred to the unlabelled data with the learned correspondence.
However, since ground truth correspondence between images is usually unavailable, the learning system must be well-supervised to avoid mode collapse and convergence failure.
To overcome this difficulty, we resort to the forward-backward consistency, which is widely used in correspondence problems, and additionally learn the backward correspondences from the warped atlases back to the original atlas.
This cycle-correspondence learning design enables a variety of extra, cycle-consistency-based supervision signals to make the training process stable, while also boost the performance.
We demonstrate the superiority of our method over both deep learning-based one-shot segmentation methods and a classical multi-atlas segmentation method via thorough experiments.
\vspace{-5pt}
\end{abstract}

\section{Introduction}
Precise segmentation of medical images delineates different anatomical structures and abnormal tissues throughout the body, which can be utilized for clinical diagnosis, treatment planning, \etc.
With sufficient well-annotated data, deep convolutional neural networks (\mbox{DCNNs}) achieved ground-breaking performance in such segmentation tasks \cite{ronneberger2015unet,milletari2016v,hesamian2019deep}.
However, obtaining 3D annotations of medical images for fully supervised training of DCNNs is labor-intensive and error-prone.
Therefore, DCNN-based segmentation methods that require only one or few examples of annotation for training are highly desirable to enable efficient development and deployment of practical solutions.

\begin{figure}[!t]
\centering
\includegraphics[width=0.9 \linewidth]{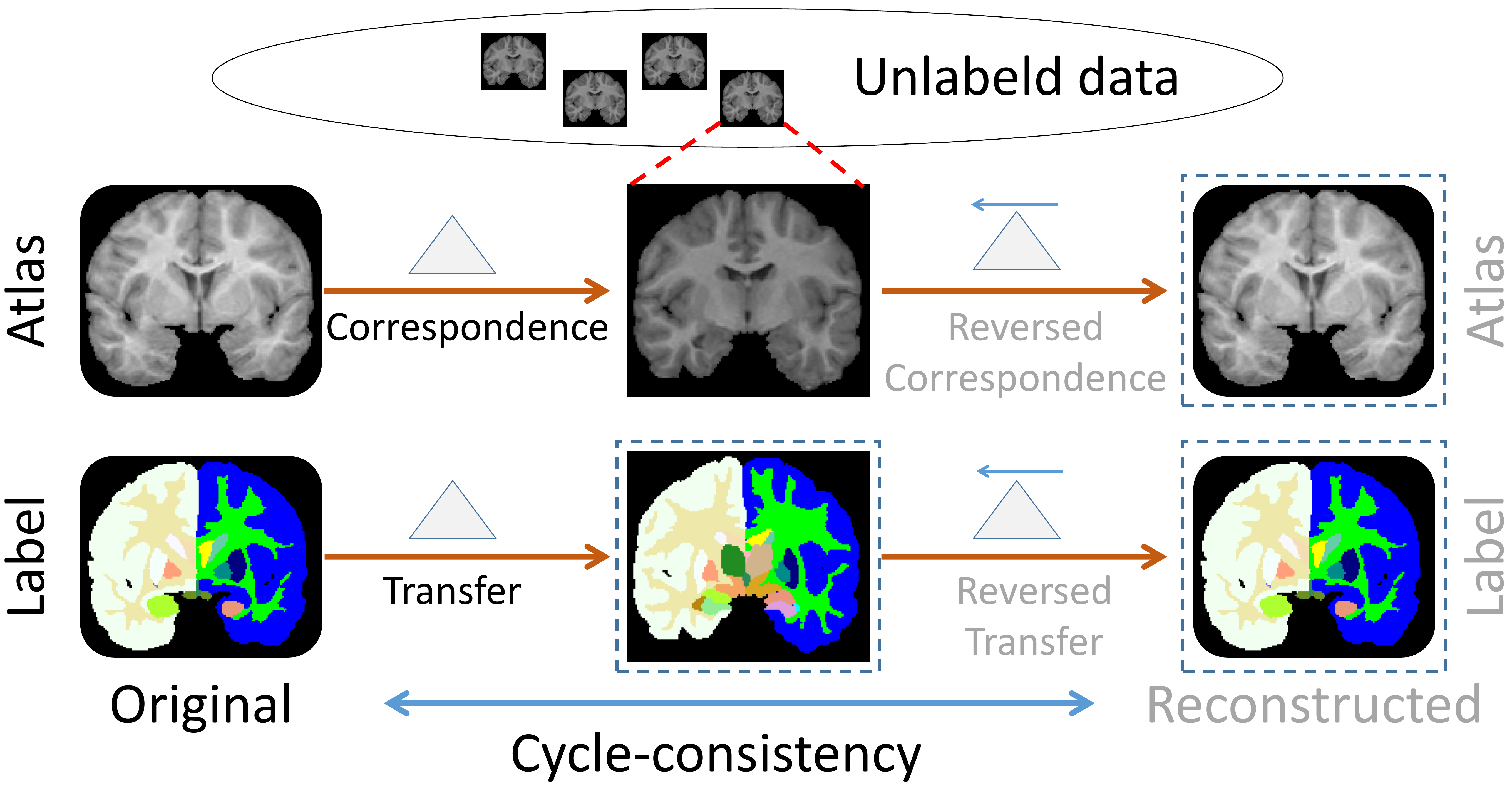}
\caption{
We tackle the one-shot medical image segmentation problem by resorting to the concept of classical atlas-based segmentation, where voxel-wise correspondence from the atlas to the unlabelled data is learned.
In addition, we innovatively learn the backward correspondences from the \emph{warped} atlases back to the original atlas to provide cycle-consistency-based supervision.
}\label{fig_demo}
\vspace{-8pt}
\end{figure}

The lack of large-scale real-world annotations is a long-standing problem in medical image segmentation.
Before the era of deep learning, a large body of literatures on medical image segmentation focused on the \textit{atlas}-based segmentation \cite{collins1995automatic,pham2000current,lorenzo2002atlas, lotjonen2010fast,wang2019patch,coupe2011patch}.
The key idea is that one or more labelled reference volumes (i.e., atlas) are non-rigidly registered \cite{collins1995automatic,pham2000current} to a target volume, or provided to learn patch-wise corresponding relationship \cite{wang2019patch,coupe2011patch} with the target volume, and then the labels of the atlases are propagated to the target volume as segmentation.
An intriguing characteristic of the atlas-based methods is that they only need one or several annotated data, naturally matching the recently rising concept of few-shot learning in deep learning.
Classical state-of-the-art (SOTA) atlas-based segmentation methods \cite{wang2019patch,coupe2011patch} rely on abundant texture features of local descriptors.
Powered by the convolutional operations repeatedly conducted in local regions, DCNNs are especially good at extracting multi-scale local semantic features.
Therefore, it is intuitive and appealing to apply DCNNs to develop advanced atlas-based methods for medical image segmentation.

Recent studies \cite{zhao2019dataaug,li2019hybrid,elmahdy2019adversarial,xu2019deepatlas,dinsdale2019spatial,yang2018neural,vakalopoulou2018atlasnet} showed that the principle of classical atlas-based segmentation could be implemented with DCNNs, and decent performance was achieved.
Among all those works, two of them are specifically related to ours in regard to one-shot learning \cite{zhao2019dataaug,xu2019deepatlas}.
In the first work, Zhao \textit{et al.} \cite{zhao2019dataaug} proposed to learn a set of spatial and appearance transformations from the atlas to unlabelled images.
By applying randomly combined spatial and appearance transformations to different unlabelled images, the model could synthesize a diverse set of labelled data.
In this sense, it provided extra labelled data for training of the segmenter.
One of the limitations of this work is that the segmentation accuracy was indirectly boosted by data augmentation, resulting in extra overhead for training the networks responsible for learning both kinds of transformations.
The second work \cite{xu2019deepatlas} proposed a framework that jointly trained two networks for image registration and segmentation, assuming that these two tasks would help each other since they were highly related.
However, in many clinical scenarios, registration is not required when segmentation is demanded.

Different from these two works,
we propose to \emph{directly} imitate the classical atlas-based segmentation with a deep learning framework, which takes both the atlas and the target image as input, and predicts the correspondence map from the former to the latter.
In this way, the segmentation label can be transferred from the atlas to the unlabelled target image with the predicted correspondence.
For efficient learning of the correspondence, we enhance the backbone of VoxelMorph \cite{balakrishnan2019voxelmorph} via the addition of a discriminator network and adversarial training \cite{goodfellow2014generative}.


Learning correspondence plays an important role in many computer vision tasks, e.g., optical flow \cite{meister2018unflow,wang2018occlusion}, tracking \cite{pan2009recurrent,kalal2010forward}, patch matching, \cite{bailer2017cnn}, registration \cite{balakrishnan2019voxelmorph,elmahdy2019adversarial,li2019hybrid,xu2019deepatlas}, and so on.
Among those inspiring works, forward-backward consistency is widely used in the correspondence problem. Specifically, our framework learns not only the forward correspondences from the atlas to unlabelled images, but also the backward correspondences from the \emph{warped} atlases back to the original atlas (see Fig. \ref{fig_demo}).
The introduction of the reverse correspondences naturally complements the full cycle of bidirectional warping, enabling extra, cycle-consistency-based supervision signals to make the learning process with only one annotation more robust and meanwhile preserve the anatomical consistency.
In addition, we impose supervision in three involved spaces, namely, the image space, the transformation space, and the label space, which has been verified effective in our experiments.

In summary, we propose a label transfer network (LT-Net) to propagate the segmentation map from the atlas to unlabelled images by learning the reversible voxel-wise correspondences. Our main contributions are as follows:

\begin{itemize}
	\item To deal with the lack of annotations, our method addresses the one-shot segmentation problem by resorting to the idea of classical atlas-based segmentation. Powered by the representation ability of the DCNN, the proposed method boosts the performance of image matching in feature space, providing anatomically meaningful correspondence for the label transfer.

	\item We extend correspondence learning to our one-shot segmentation framework in an end-to-end manner, where forward-backward cycle-consistency takes an important role to provide extra supervision in the image, transformation, and label spaces.
	\item We demonstrate the superiority of our method over both deep learning-based one-shot segmentation methods \cite{zhao2019dataaug,balakrishnan2019voxelmorph} and a classical multi-atlas segmentation method \cite{jia2012iterative} in segmenting 28 anatomical structures from a brain magnetic resonance imaging (MRI) dataset.
We also demonstrate the benefits of the cycle-consistency supervision in each individual space via ablation studies.
\end{itemize}

\section{Related Work}
\textbf{One-shot learning:}
Early works about one-shot learning mainly focused on image classification \cite{fei2003bayesian,fei2006one} based on the assumption that previously learned categories could be leveraged to help forecast a new category when very few examples are available.
Along the years, this concept has been used in various branches of machine learning and computer vision problems, such as imitation learning \cite{duan2017one,finn2017one}, object segmentation \cite{shaban2017one,caelles2017one,claudio2018,benjamin2019}, neural architecture search \cite{zhou2019,dong2019one}, and so on.
Most recently, Zhao \textit{et al.} \cite{zhao2019dataaug} developed a one-shot medical image segmentation framework based on data augmentation using learned transformations from the reference atlas to unlabelled images.
Specifically, both the spatial and appearance transformation models were learned and then utilized to synthesize additional labelled samples for data augmentation.
Our work also explores the one-shot setting for medical image segmentation to alleviate the burden of manual annotation.
The main difference is that we directly target the segmentation in our network design, and incorporate the forward-backward consistency in the framework
to ensure abundant supervision for learning.

\textbf{Atlas-based segmentation:}
Atlas-based segmentation is a classical topic in medical image analysis, evolved from single atlas-based \cite{pham2000current,zhao2019dataaug,li2019hybrid,elmahdy2019adversarial,xu2019deepatlas,dinsdale2019spatial} to sophisticated, multi-atlas-based methods \cite{klein2005mindboggle,heckemann2006automatic,wang2019patch,dinsdale2019spatial,yang2018neural,vakalopoulou2018atlasnet,jia2012iterative}.
Recently, motivated by the success of DCNNs, researchers revitalized this classical concept with deep learning models.
Using a single atlas, researchers explored this methodology in three ways: learning transformations for data augmentation \cite{zhao2019dataaug}, combining with another registration task \cite{li2019hybrid,elmahdy2019adversarial,xu2019deepatlas}, and learning a deformation field to resample an initial binary mask \cite{dinsdale2019spatial}.
Whereas for multiple atlases, recent works attempted to implement key components of multi-atlas segmentation with DCNNs, e.g., atlas selection \cite{yang2018neural}, label propagation \cite{vakalopoulou2018atlasnet}, and label fusion \cite{yang2018neural,ding2019votenet}.

Our work approaches the one-shot medical image segmentation problem via single atlas-based segmentation with a correspondence-learning generative adversarial network (GAN) framework.
It falls into the single-atlas category, which offers two advantages.
First, for complex organs, e.g., the brain, to annotate extra few samples in detail can be a considerable burden.
Second, there is no need to consider the intricate processes involved in the multi-atlas approach, such as label fusion or atlas selection.
In spite of relying on a single atlas, our proposed framework outperforms an advanced multi-atlas method \cite{jia2012iterative} using up to five atlases (cf. Section \ref{exp:atlas}).

\textbf{Correspondence in computer vision:}
Correspondence plays an important role in computer vision.
Actually, many fundamental vision problems, from optical flow \cite{meister2018unflow,wang2018occlusion} and tracking \cite{pan2009recurrent,kalal2010forward} to patch matching \cite{bailer2017cnn} and registration \cite{balakrishnan2019voxelmorph,elmahdy2019adversarial,li2019hybrid,xu2019deepatlas}, require some notion of visual correspondence \cite{wang2019learning}. Optical flow and registration can be seen as pixel/voxel-level correspondence problems, whereas tracking and patch matching can be seen as patch-level correspondence problems.
By treating atlas-based segmentation as a correspondence problem, we draw lessons from these research areas to guide the design of our framework.

\textbf{Forward-backward consistency:}
Forward-backward consistency has been widely adopted in many computer vision problems, especially in the correspondence learning problem.
For example, forward-backward consistency has been the evaluation metric \cite{kalal2010forward} as well as the measure of uncertainty \cite{pan2009recurrent} for tracking.
Recent methods on unsupervised optical flow estimation \cite{meister2018unflow,wang2018occlusion} employed forward and backward consistency to define an occluded region, which was excluded for training.
Besides, forward-backward consistency is an important building block for CycleGAN, which is the most popular framework for image-to-image translation \cite{zhu2017unpaired}.
To the best of our knowledge, our work is the first to employ cycle-consistency in one-shot atlas-based segmentation within a deep learning framework.

\section{Basic Framework for Correspondence Learning}
\label{sec:Preliminaries}
\textbf{Preliminaries:}
We first recap the basic concept of atlas-based image segmentation, where the segmentation of an unseen subject can be estimated by a registration process.
Let $(\mathit{l}, \mathit{l}_s)$ be a labelled image pair, where $\mathit{l} \in \mathbb{R}^{h\times w\times c}$ is the atlas image, $\mathit{l}_s \in \mathbb{R}^{h\times w\times c}$ is its corresponding segmentation map, and $h,w,c$ are the numbers of voxels along the coronal, sagittal, and axial directions, respectively.
In practice, input images are defined within a 3D space $\Omega \in \mathbb{R}^3$, which also applies to the unlabelled image pool ${\{\mathit{u}^{(i)}|\mathit{u}^{(i)} \in \mathbb{R}^{h\times w\times c}\}}$.
In the following, we use $\mathit{u}$ to denote an unlabelled image for an uncluttered notion.
Let
$\Delta{\mathit{p}_F}$ ($F$ standing for forward is used to differentiate the backward operations introduced in Section \ref{sec:reverse_corespondence}) denote the forward correspondence map that warps $\mathit{l}$ towards $\mathit{u}$ during the registration process.
Specifically, $\Delta{\mathit{p}_F}$ can be considered as a spatially varying function defined over $\Omega$, that maps coordinates of $\mathit{u}$ to those of $\mathit{l}$ by displacement vectors.
We use $\mathit{l} \circ \Delta{p_F}$ to denote the application of $\Delta{p_F}$ on $\mathit{l}$ (i.e., warp $\mathit{l}$ towards $\mathit{u}$ according to $\Delta{p_F}$):
\begin{equation}\label{eq:generator_F}
  \bar{\mathit{u}}=\mathit{l} \circ \Delta{p}_F,
\end{equation}
where $\circ$ is a warp operation, and $\bar{\mathit{u}}$ is the deformed atlas.
The segmentation map $\mathit{l}_s$ can be warped the same way as the atlas:
\begin{equation}\label{eq:label_trans}
  \bar{\mathit{u}}_s = \mathit{l}_s \circ \Delta{\mathit{p}}_F,
\end{equation}
where $\bar{\mathit{u}}_s$ is the synthetic segmentation of $\mathit{u}$.
If $\Delta p_F$ registers $\mathit{l}$ and $\mathit{u}$ well, $\bar{\mathit{u}}_s$ is expected to be an accurate segmentation of $\mathit{u}$.
In this sense, we treat the atlas-based segmentation as a label transfer process.

\begin{figure*}[!t]
\centering
\includegraphics[width=0.875 \textwidth]{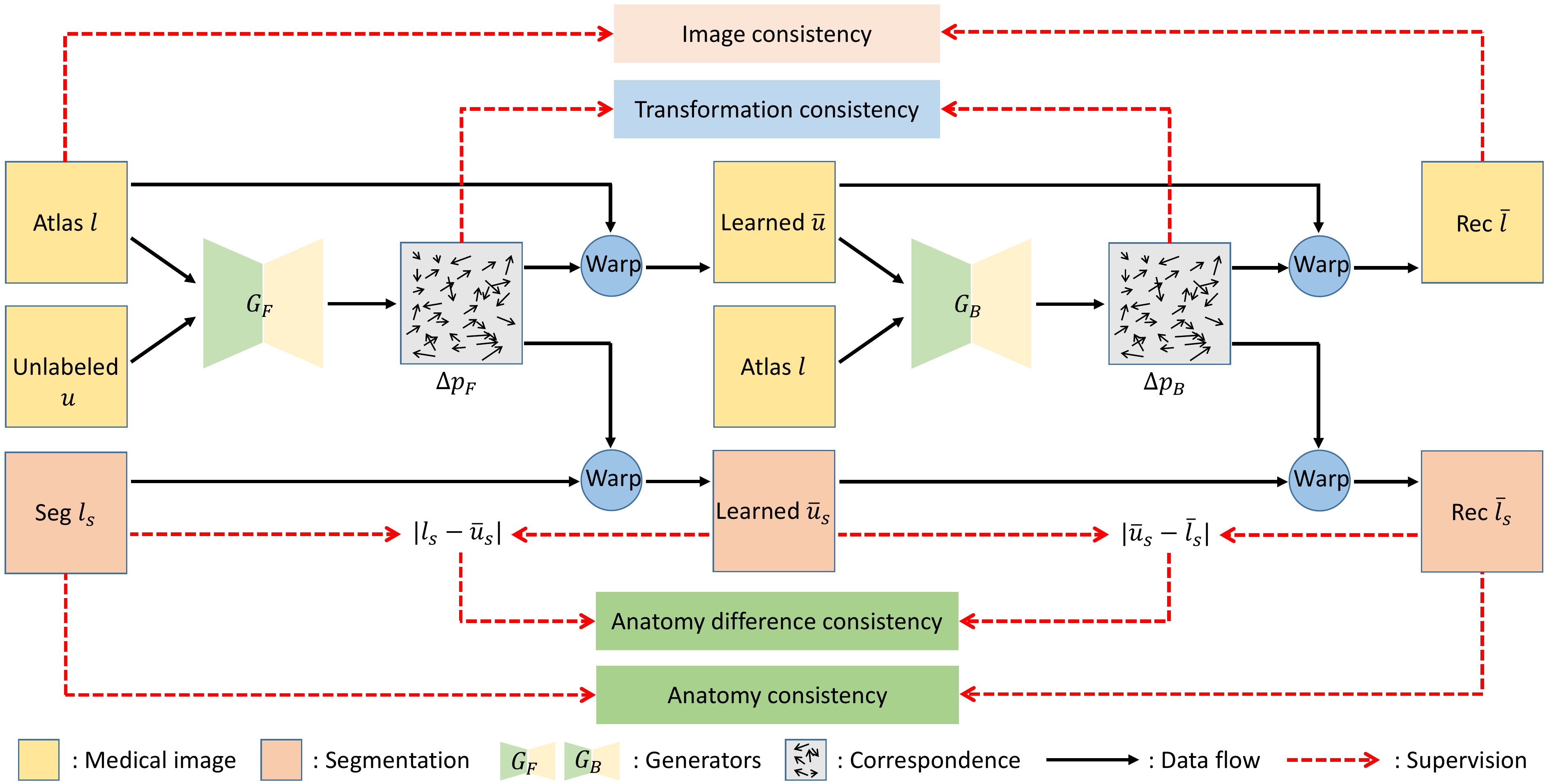}
\caption{The overview of our proposed label transfer network (LT-Net). We innovatively introduce forward-backward cycle consistency design into the atlas-based segmentation workflow. This facilitates us to explore new supervision signals, which provide more robust driving forces to guide the correspondence learning. Specifically, we adopt cycle-consistency losses in image, transformation, and label spaces and their effectiveness has been verified in the experiments.}
\label{fig_framework}
\vspace{-8pt}
\end{figure*}

\textbf{Atlas-based segmentation with deep learning:}
To model the registration function with a DCNN, a generator network $G_F$ is often adopted to match the local spatial information between $\mathit{l}$ and $\mathit{u}$, and output $\Delta p_F$.
For example, VoxelMorph \cite{balakrishnan2019voxelmorph} used a U-Net \cite{ronneberger2015unet} as $G_F$ to learn the image correspondence.
The parameters of the network are optimized by minimizing two unsupervised loss functions:
the image similarity loss $\mathcal{L}_\mathrm{sim}(\mathit{u},\bar{\mathit{u}})$ and the transformation smoothness loss $\mathcal{L}_\mathrm{smooth}(\Delta p_F)$.
To introduce robustness against global intensity variations in medical images,
we use the locally normalized cross-correlation (CC) loss \cite{xu2019deepatlas,zhao2019dataaug} for $\mathcal{L}_\mathrm{sim}$, which encourages coherence in local regions.
For $\mathcal{L}_\mathrm{smooth}$, it is formulated with first-order derivatives of $\Delta p_F$:
\begin{equation} \label{eq:smooth_loss}
\mathcal{L}_\mathrm{smooth}(\Delta p_F)= \sum_{ \mathit{t} \in \Omega}\Vert \nabla{(\Delta{p_F(t)})} \Vert_{2},
\end{equation}
where $t\in\Omega$ iterates over all spatial locations in $\Delta p_F$, and we approximate $\Vert \nabla{(\Delta{\mathit{p}(t)})} \Vert_{2}$ with spatial gradient differences between neighboring voxels along $x,y,z$ directions \cite{balakrishnan2019voxelmorph}.
Minimizing $\mathcal{L}_\mathrm{sim}$ encourages $\bar{\mathit{u}}$ to approximate $\mathit{u}$, whereas minimizing $\mathcal{L}_\mathrm{smooth}$ regularizes $\Delta p_F$ to be smooth.
In addition, smoothness regularization can be considered as a strategy to alleviate the overfitting problem while encoding the anatomical priori.


\textbf{Auxiliary GAN loss:}
Besides the two basic losses used by VoxelMorph, we introduce a GAN \cite{goodfellow2014generative} into our basic framework to offer additional supervision.
The GAN subnet in our framework comprises $G_F$ and a discriminator network $D$.
A vanilla GAN would make the discriminator $D$ differentiate $\Delta{\mathit{p}_F}$ from the true underlying correspondence map. In practice, however, it is usually infeasible to obtain the true correspondence between a pair of clinical images. 
Instead, we make $D$ distinguish the synthetic image $\bar{\mathit{u}}$ from $\mathit{u}$.
In this sense, $\bar{\mathit{u}}$ serves as a delegate of $\Delta{\mathit{p}_F}$, and $G_F$ is trained to generate $\Delta{\mathit{p}_F}$ that can be used to synthesize $\bar{\mathit{u}}$ authentically enough to confuse $D$;
meanwhile, $D$ becomes more skilled at flagging synthesized images.
This delegation strategy provides indirect supervision to $G_F$ and $\Delta p_F$, and allows the networks to be trained end-to-end with a large number of unlabelled images.
Consequently, the image adversarial loss $\mathcal{L}_\mathrm{GAN}$ is defined as
\begin{equation} \label{eq:gan_loss}
\begin{split}
\mathcal{L}_\mathrm{GAN}&(\mathit{l}, \mathit{u}, \bar{\mathit{u}}) = \mathbb{E}_{\mathit{u} \sim p_d(\mathit{u})}
[\Vert D(\mathit{u}) \Vert_2] \\
&+ \mathbb{E}_{\mathit{l} \sim p_d(\mathit{l}),\mathit{u} \sim p_d(\mathit{u})}
[\Vert D(\bar{\mathit{u}}) - \mathbf{1} \Vert_2],
\end{split}
\end{equation}
where $G_F$ and $D$ are trained alternatively to compete in a two-player min-max game with the objective function $\min_{G_F} \max_{D} \mathcal{L}_\mathrm{GAN}(G_F, D)$.
\footnote{Our basic framework for correspondence learning is illustrated and explained in more details in the Appendix.}

\section{Learning Reversible Cycle Correspondence}
\label{sec:reverse_corespondence}
In the previous section, we introduce our baseline method for atlas-based one-shot segmentation.
Our proposed framework is built on this baseline and adds a cycle consistency constraint to further boost the segmentation performance. We name the proposed framework as label transfer network (LT-Net).
Unlike previous works \cite{zhao2019dataaug,xu2019deepatlas} that only learned the forward correspondences from the atlas to unlabelled images, we in addition learn the backward correspondences from the \emph{warped} atlases back to the original atlas.
As far as the authors are aware of, this work is the first attempt that utilizes the cycle correspondence for one-shot (atlas-based) segmentation with deep learning.
Specifically,
we propose a backward correspondence learning path:
$\Delta p_B = G_B(\bar{\mathit{u}}, \mathit{l})$, where $\Delta p_B$ is the backward correspondence map, and $G_B$ is the backward generator (see $\Delta p_B$ and $G_B$ in Fig. \ref{fig_framework}).
With the newly added $\Delta p_B$, we can \emph{revert} the synthetic image $\bar{\mathit{u}}$ to reconstruct the atlas using the warp operation:
\begin{equation}\label{eq:generator_B}
  \bar{\mathit{l}} = \bar{\mathit{u}} \circ \Delta{\mathit{p}}_B,
\end{equation}
and we call $\bar{\mathit{l}}$ the reconstructed atlas.
Accompanied with the backward learning path, a straightforward addition to the network's supervision is to impose transformation smoothness loss on $\Delta p_B$ as well.
Hence, our complete transformation smoothness loss becomes
\begin{equation}\label{eq:complete_smooth_loss}
  \mathcal{L}_\mathrm{smooth} = \mathcal{L}_\mathrm{smooth}(\Delta p_F) + \mathcal{L}_\mathrm{smooth}(\Delta p_B).
\end{equation}

More importantly, the completion of the correspondence cycle enables a variety of supervision signals to boost the performance upon unidirectional correspondence learning.
Concretely, we propose four novel, cycle-consistency-driven supervision losses (cf. the supervision blocks in Fig. \ref{fig_framework}) in three spaces, namely, the image space, the transformation space, and the label space.
These supervision losses are all devised by straightforward intuitions, as described below.
\begin{itemize}
  \item In the image space, the reconstructed and original atlas images ($\bar{\mathit{l}}$ and $\mathit{l}$) should be the same (the \emph{image consistency}).
  \item In the transformation space, conceptually the forward and backward warpings should be the inverse function of each other, so that the atlas warped toward the unlabelled image can be warped back to what it originally is (the \emph{transformation consistency}).
  \item Lastly, in the label space, the true segmentation $\mathit{l}_s$ and the reconstructed segmentation $\bar{\mathit{l}}_s$ should be the same (the \emph{anatomy consistency}).
      In addition, they must differ from the synthetic segmentation $\bar{\mathit{u}}_s$ in the same way (the \emph{anatomy difference consistency}).
\end{itemize}
Despite being conceptually simple, the comprehensive inclusion and combination of these supervision signals in our framework are proved to be effective in the experiments---our LT-Net outperforms the current SOTA by significant margins, and the ablation studies demonstrate benefits of the supervision in individual spaces. In the following, we describe each loss in detail.

\textbf{Cycle-consistency supervision in image space:}
Enabled by our novel forward-backward cycle correspondence learning framework, we can revert the synthetic image $\bar{\mathit{u}}$ to reconstruct the atlas. We employ an L1 loss to enforce the consistency between the true atlas and the reconstructed one, which is defined as
\begin{equation} \label{eq:cycle_loss}
\mathcal{L}_\mathrm{cyc}(\mathit{l}, \bar{\mathit{l}}) = \mathbb{E}_{\mathit{l} \sim p_d(\mathit{l})} [\Vert \bar{\mathit{l}} - \mathit{l} \Vert_1].
\end{equation}

\textbf{Cycle-consistency supervision in transformation space:}
In terms of forward-backward consistency, the correspondences should be reversible, meaning that a voxel warped from one position to another in the forward path should be warped back to its original position in the backward path.
Therefore, we define a transformation consistency loss to enforce this constraint as
\begin{equation} \label{eq:transformation_loss1}
\footnotesize
\begin{split}
&\mathcal{L}_\mathrm{trans}(\Delta{\mathit{p}_{F}}, \Delta{\mathit{p}_{B}})=\sum_{t \in \Omega} \rho(\Delta{\mathit{p}}_{F}(t) + \Delta{\mathit{p}_{B}(t+ \Delta{\mathit{p}}_{F}(t))}),
\end{split}
\end{equation}
where $\rho(\mathit{x})=(\mathit{x}^2+\epsilon^2)^\gamma$ is a robust generalized Charbonnier penalty function \cite{sun2014quantitative} and widely used as a photometric loss in optical flow estimation
\cite{meister2018unflow,wang2018occlusion}.
In this work, we use the same setting with $\epsilon=0.001,\gamma=0.45$ as \cite{meister2018unflow}.

\textbf{Cycle-consistency supervision in label space:}
In many applications, matching the images solely based on intensity is under-constrained and may lead to wrong correspondences. The corresponding anatomical structure may shift or twist away from one position to another, as long as the warped and target images appear similar. Enforcing smoothness constraint on the correspondence map (as in VoxelMorph \cite{balakrishnan2019voxelmorph}) is a common way of alleviating this problem. In this work, we further explore driving forces in the label space to guide the correspondence learning towards an anatomically meaningful direction.

When considering supervision signal in the label space, an anatomy cycle-consistency constraint naturally comes up within our framework.
Let $\bar{\mathit{l}}_s = \bar{\mathit{u}}_s \circ \Delta{\mathit{p}}_B$ denote the reconstructed segmentation map of $\bar{\mathit{l}}$.
To model the dissimilarity between $\bar{\mathit{l}}_s$ and the original segmentation map $\mathit{l}_s$, a Dice loss \cite{milletari2016v} is adopted which is defined as
\begin{equation} \label{eq:dice_loss}
\mathcal{L}_\mathrm{anatomy\_cyc}(\mathit{l}_s, \bar{\mathit{l}}_s) = 1-\frac{2 \sum\limits_{t \in \Omega} \mathit{l}_s(t) \bar{\mathit{l}}_s(t)}{\sum\limits_{t \in \Omega} \mathit{l}_s^2(t) + \sum\limits_{t \in \Omega} \bar{\mathit{l}}_s^2(t)}.
\end{equation}
Since our target is to learn the correspondence which can be used to transfer the segmentation map of the atlas to each of the unlabelled images, we also propose an anatomy difference consistency loss to indirectly regularize quality of the synthetic segmentation map $\bar{\mathit{u}}_s$.
As aforementioned, this loss is based on a simple intuition: the anatomy differences between the atlas and the unlabelled image in the forward and backward paths should be cyclically consistent in the label space. The loss is thus formulated as
\begin{equation} \label{eq:transformation_loss2}
\begin{split}
\mathcal{L}_\mathrm{diff\_cyc}&(\mathit{l}_s, \bar{\mathit{u}}_s, \bar{\mathit{l}}_s) = \\
&\sum_{t \in \Omega} \rho(|\mathit{l}_s(t) - \bar{\mathit{u}}_s(t)| - |\bar{\mathit{u}}_s(t) - \bar{\mathit{l}}_s(t)|).
\end{split}
\end{equation}


\section{Optimization Objective and Implementation}
Given the definitions of the supervision signals above, our complete objective for optimization is defined as
\begin{equation} \label{eq:objective}
\begin{split}
\mathcal{L} = & \mathcal{L}_\mathrm{GAN} + \mathcal{L}_\mathrm{sim} + \lambda_1 \mathcal{L}_\mathrm{cyc} + \lambda_2 (\mathcal{L}_\mathrm{anatomy\_cyc} \\
&+ \mathcal{L}_\mathrm{smooth} + \mathcal{L}_\mathrm{trans} + \mathcal{L}_\mathrm{diff\_cyc}),\\
\end{split}
\end{equation}
where $\lambda_1$ and $\lambda_2$ are the weights to balance the importance of the different losses.
We use the same weight for the last four losses in Eq. (\ref{eq:objective}), since they are comparable in magnitude and we find the results insensitive to their relative weights in our primitive experiments.
We set $\lambda_1=10$ following CycleGAN \cite{zhu2017unpaired}, and consequently set $\lambda_2=3$ to make the corresponding loss values at the same level as $\mathcal{L}_\mathrm{cyc}$.
The supervision signals in the image, transformation, and label spaces affect each other and restrict each other, pushing the learning system towards an anatomically meaningful direction.

We implement all models using Keras~\cite{chollet2015keras} with a TensorFlow~\cite{abadi2016tensorflow} backend.
For the generator networks in both the forward and backward paths, we adopt the same 3D U-Net architecture as VoxelMorph ~\cite{balakrishnan2019voxelmorph} for a fair comparison later.
For the discriminator network, we use an extended 3D version of PatchGAN~\cite{isola2017image} to determine whether an image patch is real or synthesized.
All networks are optimized from scratch using the Adam solver~\cite{kingma2014adam}. The learning rate is initialized to $0.0002$ and remains unchanged during the training process.
Each mini-batch processes a pair of volumes (one atlas and one unlabelled image) per GPU while running two Tesla P40 GPUs in parallel.
During testing, the forward correspondence map $\Delta p_F$ from the atlas to a test unlabelled image $\mathit{u}^{(i)}$ is predicted by $G_F$, then the segmentation map for $\mathit{u}^{(i)}$ is produced with Eq. (\ref{eq:label_trans}).

\section{Experiments}
We demonstrate the superiority of our LT-Net on the task of brain MRI segmentation.
Above all, the effectiveness of the cycle correspondence learning framework is evaluated (Section \ref{exp:for_back}).
As aforementioned, the forward-backward consistency is a classical constraint in correspondence problems.
By introducing a backward correspondence path, extra meaningful supervision signals can be exploited to drive the learning process towards a more robust and anatomically meaningful direction.
Hence, within the cycle correspondence framework, we subsequently examine the effects of the several newly proposed cycle consistency losses with ablation studies: the transformation consistency loss in the transformation space, the anatomy consistency and difference consistency losses in the label space, and the combination of the losses from both spaces (Section \ref{exp:sup_sig}).
Next, we compare our method with a classical multi-atlas method (Section \ref{exp:atlas}), demonstrating that the traditional idea of atlas-based segmentation in computer vision can be further boosted using deep learning.
Finally, we compare with a SOTA method for one-shot medical image segmentation, demonstrating the superiority of our framework to other DCNN-based methods (Section \ref{exp:fully}).
Examples of the synthesized images and warped segmentation maps for unlabelled images are also presented for visual evaluation.

\subsection{Dataset and Evaluation Metric}
\textbf{Dataset:} We use a publicly available dataset from the Child and Adolescent NeuroDevelopment Initiative (CANDI) at the University of Massachusetts Medical School~\cite{kennedy2011candishare}.
The dataset comprises 103 T1-weighted MRI scans (57 males and 46 females) with anatomic segmentation labels. The subjects come from four diagnostic groups: healthy controls, schizophrenia spectrum, bipolar disorder with psychosis, and bipolar disorder without psychosis.
We use 28 anatomical structures (tabulated in the Appendix) that were used in VoxelMorph \cite{balakrishnan2019voxelmorph}.
The volume size ranges from $256 \times 256 \times 128$ to $256 \times 256 \times 158$ voxels.
For computation efficiency, we crop a $160 \times 160 \times 128$ region around the center of the brain, which is large enough to contain the whole brain.
We randomly select 20 volumes as test data, and use the others for training.
Among the training data, the volume which is most similar to the anatomical average is selected as the only annotated atlas (the same strategy as adopted in VoxelMorph \cite{balakrishnan2019voxelmorph}).


\textbf{Evaluation metric:} We use the Dice similarity coefficient~\cite{dice1945measures} to evaluate the segmentation accuracy of each model, which measures the overlap between manual annotations and predicted results.

\subsection{Effectiveness of Forward-backward Consistency} \label{exp:for_back}

\begin{table}[!t] \caption {Mean Dice scores (\%) (with standard deviations in parentheses) for VoxelMorph \cite{balakrishnan2019voxelmorph}, and its extended versions which gradually incorporate the image adversarial loss $\mathcal{L}_\mathrm{GAN}$ and cycle-consistency loss $\mathcal{L}_\mathrm{cyc}$. Min and Max represent the minimum and maximum Dice scores (\%) in the test dataset.} \label{arch_compare}
	\centering\scalebox{.9}{
	\begin{threeparttable}

	\begin{tabular}{l|lll}
			& {Mean (std)} & {Min} & {Max} \\
			\hline
            VoxelMorph &76.0 (9.7) &61.7 &80.1\\
            + $\mathcal{L}_\mathrm{GAN}$ & 79.0 (3.1) & 72.7 & 81.9 \\
			+ $\mathcal{L}_\mathrm{GAN}$ + $\mathcal{L}_\mathrm{cyc}$ & \textbf{79.2} (\textbf{2.8}) & 72.7 & 82.1\\
		\end{tabular}
	\end{threeparttable}}
\vspace{-5pt}
\end{table}

We adapt VoxelMorph \cite{balakrishnan2019voxelmorph}---the SOTA DCNN registration model---for our problem setting and use it as the initial performance baseline.
Specifically, we train it as a single-atlas model to learn the forward correspondence, and warp the atlas's segmentation map according to the learned correspondence for each unlabelled image.
As introduced in Section \ref{sec:Preliminaries}, our basic framework adopts the same backbone as VoxelMorph for forward correspondence learning, but adds a GAN for additional supervision.
Then, built on top of the basic framework, our LT-Net introduces a backward correspondence learning path to form a complete cycle correspondence framework.
Enabled by the cyclic structure, we further add an image cycle-consistency loss $\mathcal{L}_\mathrm{cyc}$ between the atlas and the reconstructed one.
For detailed comparisons, we first add $\mathcal{L}_\mathrm{GAN}$ alone, and then $\mathcal{L}_\mathrm{cyc}$ together.

The quantitative comparison results are shown in Table~\ref{arch_compare}.
The results show that $3.0\%$ and $3.2\%$ improvements are achieved when gradually adding $\mathcal{L}_\mathrm{GAN}$ and $\mathcal{L}_\mathrm{cyc}$.
This indicates that $\mathcal{L}_\mathrm{GAN}$ can boost the performance for our correspondence learning problem, which is in accordance with the practical experience that image adversarial losses usually perform well in image-to-image translation tasks.
It is worth noting that $\mathcal{L}_\mathrm{cyc}$ does not bring substantial further improvement upon the GAN setting.
However, as we have mentioned earlier and will experimentally show next, the benefit of the cycle design is that it enables us to incorporate extra supervision signals to the learning framework, which can further improve the performance.
Next, we treat the VoxelMorph backbone plus $\mathcal{L}_\mathrm{GAN}$ and $\mathcal{L}_\mathrm{cyc}$ as a new baseline, and design experiments to examine the effectiveness of the newly proposed supervision signals.

\subsection{Ablation Study on the Supervision Signals} \label{exp:sup_sig}

\begin{table}
	\caption{Ablation study on the newly proposed supervision signals. We show the mean Dice scores (\%) with standard deviations. Besides, Min and Max represent the minimum and maximum Dice scores (\%) in the test dataset.}\label{tab:ablation_signals}
	\centering\scalebox{0.9}{
	\begin{threeparttable}
	\begin{tabular}{l|lll}
		& {Mean (std)} & {Min} & {Max} \\
		\hline
		Baseline &79.2 (2.8) &72.7 &82.1\\
		+ $\mathcal{L}_\mathrm{trans}$ & 80.9 (2.7) &73.6 & 83.2\\
		+ $\mathcal{L}_\mathrm{anatomy\_cyc}$ &80.5 (2.5) &74.2 &83.1\\
		\footnotesize{+ $\mathcal{L}_\mathrm{trans}$ + $\mathcal{L}_\mathrm{anatomy\_cyc}$} &81.4 (2.6) &74.4 &83.8\\
		\footnotesize{+$\mathcal{L}_\mathrm{trans}$ + $\mathcal{L}_\mathrm{anatomy\_cyc}$} &\multirow{2}{*}{\textbf{82.3} (\textbf{2.5})} &\multirow{2}{*}{75.6} &\multirow{2}{*}{84.2}\\
				\qquad\footnotesize{+ $\mathcal{L}_\mathrm{diff\_cyc}$}
	\end{tabular}
	\end{threeparttable}}
\vspace{-5pt}
\end{table}

\textbf{Cycle-consistency in transformation space:}
The correspondence learned from the atlas to the unlabelled image is used to synthesize $\bar{\mathit{u}}$ by warping the atlas in the forward path, whereas in the backward path another correspondence is learned from $\bar{\mathit{u}}$ back to the atlas.
The forward and backward correspondences should be cycle-consistent.
We conduct an ablation study with respect to the transformation consistency loss $\mathcal{L}_\mathrm{trans}$ and show the results in Table \ref{tab:ablation_signals}.
From the table, we can observe that $\mathcal{L}_\mathrm{trans}$ brings a $1.7\%$ improvement compared to the baseline.
This may imply that intensity matching at the image level---despite the cycle correspondence setting---is not enough to prevent the overfitting by DCNNs, and the introduction of supervision in other spaces (e.g., the label space) has the potential for further improvement in performance.

\begin{table}[!t] \caption {Comparison of our LT-Net with a multi-atlas method MABMIS \cite{jia2012iterative} using increasing numbers of atlases. We show the mean Dice scores (\%) with standard deviations. Besides, Min and Max represent the minimum and maximum Dice scores (\%) in the test dataset.} \label{tab:compare_atlas}
	\centering\scalebox{0.9}{
	\begin{threeparttable}

	\begin{tabular}{c|c|ccc}
            & {No. of Atlases} & {Mean (std)} & {Min} & {Max}\\
			\hline
			MABMIS & 2 & 70.4 (4.4) & 63.0 & 75.9\\
			MABMIS & 3 & 74.5 (5.2) & 63.2 & 80.4\\
			MABMIS & 4 & 77.8 (3.9) & 72.0 & 82.5\\
			MABMIS & 5 & 81.1 (3.6) & 76.0 & 85.6\\
            LT-Net & 1 & \textbf{82.3} (\textbf{2.5}) & 75.6 & 84.2\\
		\end{tabular}
	\end{threeparttable}}
\vspace{-5pt}
\end{table}

\textbf{Cycle-consistency in label space:}
With the forward correspondence, the segmentation map of the atlas can be warped to synthesize the segmentation map for each unlabelled image.
Inversely, the synthetic segmentation map can be warped back to restore the segmentation map of the atlas using the backward correspondence.
Ideally, the segmentation maps of the atlas before and after the dual warping should be the same, and we enforce this constraint with the anatomy consistency loss $\mathcal{L}_\mathrm{anatomy\_cyc}$.
Table \ref{tab:ablation_signals} quantitatively displays the effect of this supervision.
We can observe that $\mathcal{L}_\mathrm{anatomy\_cyc}$ brings a $1.3\%$ improvement when compared with the baseline, and an extra $0.9\%$ improvement when further combined with the transformation consistency loss.
As expected, $\mathcal{L}_\mathrm{anatomy\_cyc}$ boosts the performance, since it can ensure the integrity and internal coherence of the anatomical structure.

The anatomy cycle-consistency loss does not consider the middle-cycle synthesized segmentation map $\bar{\mathit{u}}_s$ for each unlabelled image, which, however, is the ultimate goal of our LT-Net.
To place more emphasis on $\bar{\mathit{u}}_s$, the anatomy difference consistency loss $\mathcal{L}_\mathrm{diff\_cyc}$ is proposed in the label space to regularize the differences between the segmentation maps of the atlas and that of the unlabelled image.
The results in Table \ref{tab:ablation_signals} show that by indirectly regularizing the segmentation maps of the unlabelled images, we achieve a $0.9\%$ further improvement.

\begin{table*}[!t] \caption {Segmentation accuracy (mean Dice scores, \%) of VoxelMorph \cite{balakrishnan2019voxelmorph}, DataAug \cite{zhao2019dataaug}, U-Net \cite{ronneberger2015unet} and our proposed LT-Net across various brain structures. Labels consisting of left and right structures are combined (\textit{e.g.}, hippocampus).
Abbreviations: white matter (WM), cortex (CX), ventricle (Vent), and cerebrospinal fluid (CSF).} \label{tab:fully_sup_structures}
	\centering
	\footnotesize
	\setlength{\tabcolsep}{0.2mm}{
	\begin{tabular}{p{2.5cm}|p{0.8cm}p{0.8cm}p{0.8cm}p{0.8cm}p{0.8cm}p{0.8cm}p{0.8cm}p{0.8cm}p{0.8cm}p{0.8cm}p{0.8cm}p{0.8cm}p{0.8cm}p{0.8cm}p{0.8cm}p{0.8cm}}
			&\rotatebox{60}{Cerebral-WM} &\rotatebox{60}{Cerebral-CX} &\rotatebox{60}{Lateral-Vent} &\rotatebox{60}{Cerebellum-WM} &\rotatebox{60}{Cerebellum-CX} &\rotatebox{60}{Thalamus-Proper} &\rotatebox{60}{Caudate}  &\rotatebox{60}{Putamen} &\rotatebox{60}{Pallidum} &\rotatebox{60}{3rd-Vent} &\rotatebox{60}{4th-Vent} &\rotatebox{60}{Brain-Stem} &\rotatebox{60}{Hippocampus} &\rotatebox{60}{Amygdala} &\rotatebox{60}{CSF} &\rotatebox{60}{VentralDC}\\
			\hline
			VoxelMorph &81.7 &87.1 &76.7 &74.0 &86.3 &84.7 &79.6 &83.3 &74.0 &62.9 &69.8 &87.1 &59.2 &66.5 &50.9 &76.6 \\
			DataAug &90.5 &93.3 &87.5 &82.3 &93.2 &87.3 &81.3 &82.8 &73.7 &66.3 &72.2 &90.5 &72.5 &69.2 &63.3 &80.3 \\
            LT-Net &85.8 &90.9 &83.1 &80.0 &91.6 &87.9 &85.5 &88.4 &80.5 &68.4 &79.7 &92.4 &71.6 &71.6 &67.1 &82.3\\
            U-Net (upper bound) &92.0 &93.1 &91.8 &87.9 &93.1 &90.6 &88.1 \qquad\quad &88.7 &82.5 &79.0 &84.9 &92.2 &80.3 &75.3 &69.8 &85.9\\
	\end{tabular}}
\vspace{-5pt}
\end{table*}

\subsection{Comparison with a Classical Multi-atlas Method} \label{exp:atlas}
Traditional multi-atlas methods once achieved SOTA results for atlas-based segmentation.
We compare our LT-Net with MABMIS \cite{jia2012iterative}, which consists of a tree-based groupwise registration method and an iterative groupwise segmentation method.
The results are shown in Table \ref{tab:compare_atlas}.
We can observe that our method using only one atlas outperforms MABMIS using up to five atlases.
In addition, classical multi-atlas segmentation is notorious for being time-consuming.
While MABMIS requires $\sim$14 minutes to segment one case with an Intel\textsuperscript{\textregistered} Core i3-4150 CPU (using two atlases), our LT-Net only needs $\sim$4 seconds with a single Tesla P40 GPU.

\subsection{Comparison with SOTA Methods} \label{exp:fully}
Besides VoxelMorph, we also compare our proposed LT-Net with DataAug \cite{zhao2019dataaug}, a SOTA method for one-shot medical image segmentation relying on registration-based data augmentation.
In addition, we train a fully supervised U-Net \cite{ronneberger2015unet} using a labelled training pool of 83 subjects, which is served as the upper bound for the one-shot methods.
The results are shown in Table \ref{tab:upper_bound}.
Using only one annotated data for training, our framework achieves $95.1\%$ of the upper bound on the mean Dice score, yet with an apparently lower standard deviation.
Besides, we can observe that our LT-Net outperforms both VoxelMorph and DataAug by margins of $6.3\%$ and $1.9\%$, respectively.
Table \ref{tab:fully_sup_structures} shows the segmentation accuracy across various brain structures.

\begin{figure}[!t]
\centering
\includegraphics[width=0.85 \linewidth]{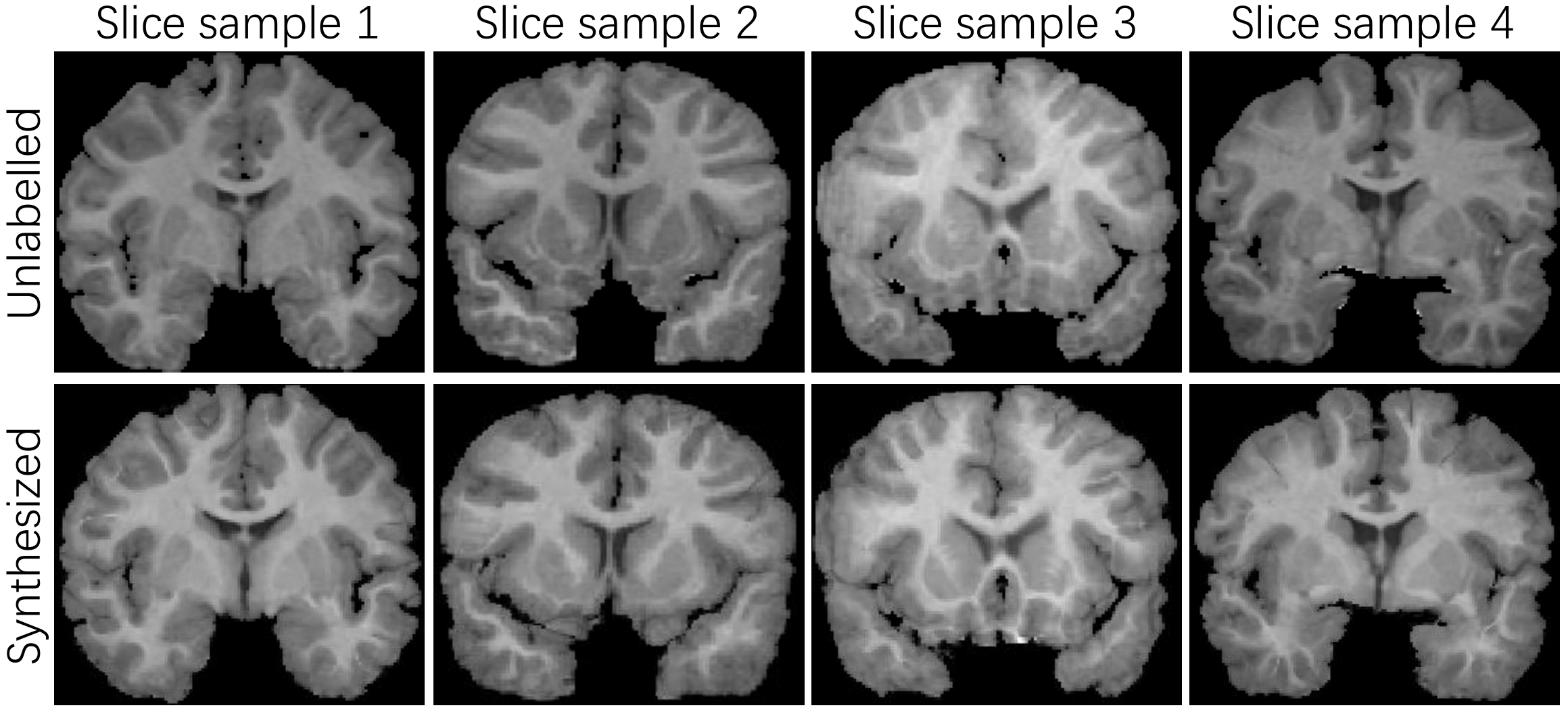}
\caption{Example coronal MR slices from the synthetic images obtained by warping the atlas with the learned forward correspondences. Each column is a different patient.}
\label{fig_reconstructed}
\vspace{-3mm}
\end{figure}

\begin{figure}[!t]
\centering
\includegraphics[width=0.95 \linewidth]{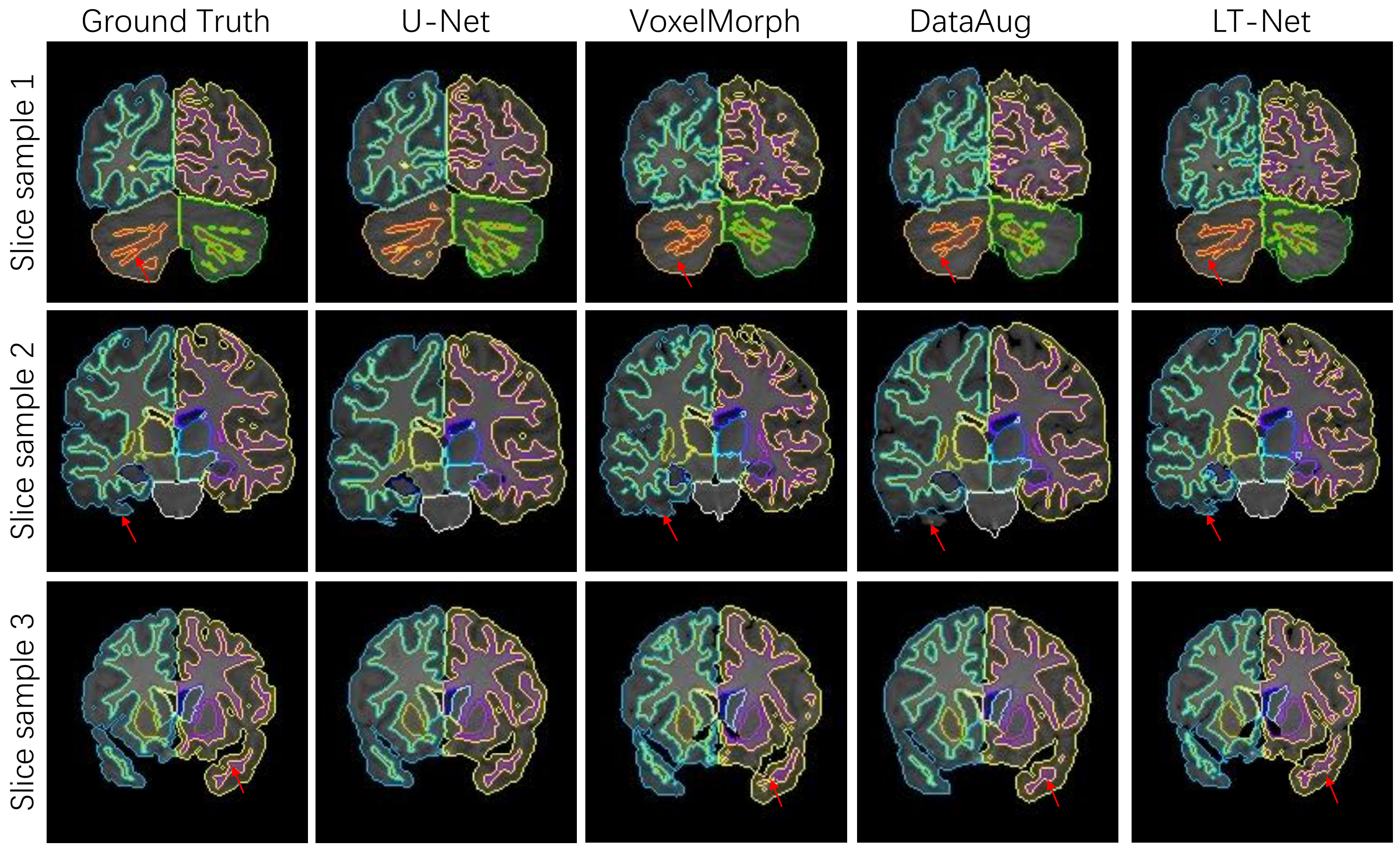}
\caption{Example coronal MR slices of brain structure annotations and segmentation maps predicted by VoxelMorph \cite{balakrishnan2019voxelmorph}, DataAug \cite{zhao2019dataaug},  U-Net \cite{ronneberger2015unet} and our LT-Net. Each row is a different patient.
Red arrows point to the flaws (best viewed zoomed-in) in the predictions made by the other one-shot methods, as compared to those by LT-Net.}
\label{fig_segmentation}
\vspace{-5pt}
\end{figure}

We visualize some example slices of the synthetic volumes $\bar{\mathit{u}}$ from different patients in Fig. \ref{fig_reconstructed}.
We observe that the synthesized images are close to the unlabelled images in terms of the anatomical structures.
In addition, Fig.~\ref{fig_segmentation} shows some example slices of brain structure annotations and segmentation maps predicted by U-Net, VoxelMorph, DataAug, and our proposed LT-Net.
Compared to the other two one-shot methods, LT-Net predicts brain structures in a way that is more anatomically meaningful.

\begin{table}[!t] \caption {Comparison of our LT-Net with VoxelMorph \cite{balakrishnan2019voxelmorph}, DataAug \cite{zhao2019dataaug} and fully supervised U-Net \cite{ronneberger2015unet}. We show the mean Dice scores (\%) with standard deviations. Besides, Min and Max represent the minimum and maximum Dice scores (\%) in the test dataset.} \label{tab:upper_bound}
    \centering\scalebox{0.9}{
    \setlength{\tabcolsep}{4mm}{
	\begin{threeparttable}
	\begin{tabular}{l|lll}
            & {Mean (std)} & {Min} & {Max}\\
			\hline
			VoxelMorph &76.0 (9.7) &61.7 &80.1\\
			DataAug &80.4 (4.3) &73.8 &84.0\\
            LT-Net &82.3 (2.5) &75.6 &84.2\\
            U-Net (upper bound) &86.5 (6.3) &83.7 &89.2\\
		\end{tabular}
	\end{threeparttable}}}
\vspace{-5pt}
\end{table}

\section{Conclusion}


In this study, we traced back to two classical ideas---atlas-based segmentation and correspondence---in computer vision and applied them to one-shot medical image segmentation with DCNNs.
First, we bridged the conceptual gap between atlas-based segmentation and the generic idea of one-shot segmentation.
This provided us with some critical thinkings for the design of our deep network.
Second, we adopted the forward-backward consistency strategy from other correspondence problems, which subsequently enabled the design of a few novel supervision signals in three involved spaces (namely, the image space, the transformation space, and the label space) to make the learning well-supervised and effectively-guided.
We hope this work would inspire the future development of one-shot learning for medical image segmentation in the era of deep learning.

\section*{Acknowledgment}
This work was supported by the National Natural Science Foundation of China (Grant No. 61671399), the Fundamental Research Funds for the Central Universities (Grant No. 20720190012), the Key Area Research and Development Program of Guangdong Province, China (Grant No. 2018B010111001) and the Science and Technology Program of Shenzhen, China (No. ZDSYS201802021814180).

\renewcommand\thesection{\Alph{section}}
\setcounter{section}{0}
\setcounter{figure}{0}
\setcounter{table}{0}
\renewcommand{\thefigure}{A\arabic{figure}}
\renewcommand{\thetable}{A\arabic{table}}
\section{Appendix}

\begin{figure*}[t]
\centering
\includegraphics[width=0.75 \textwidth]{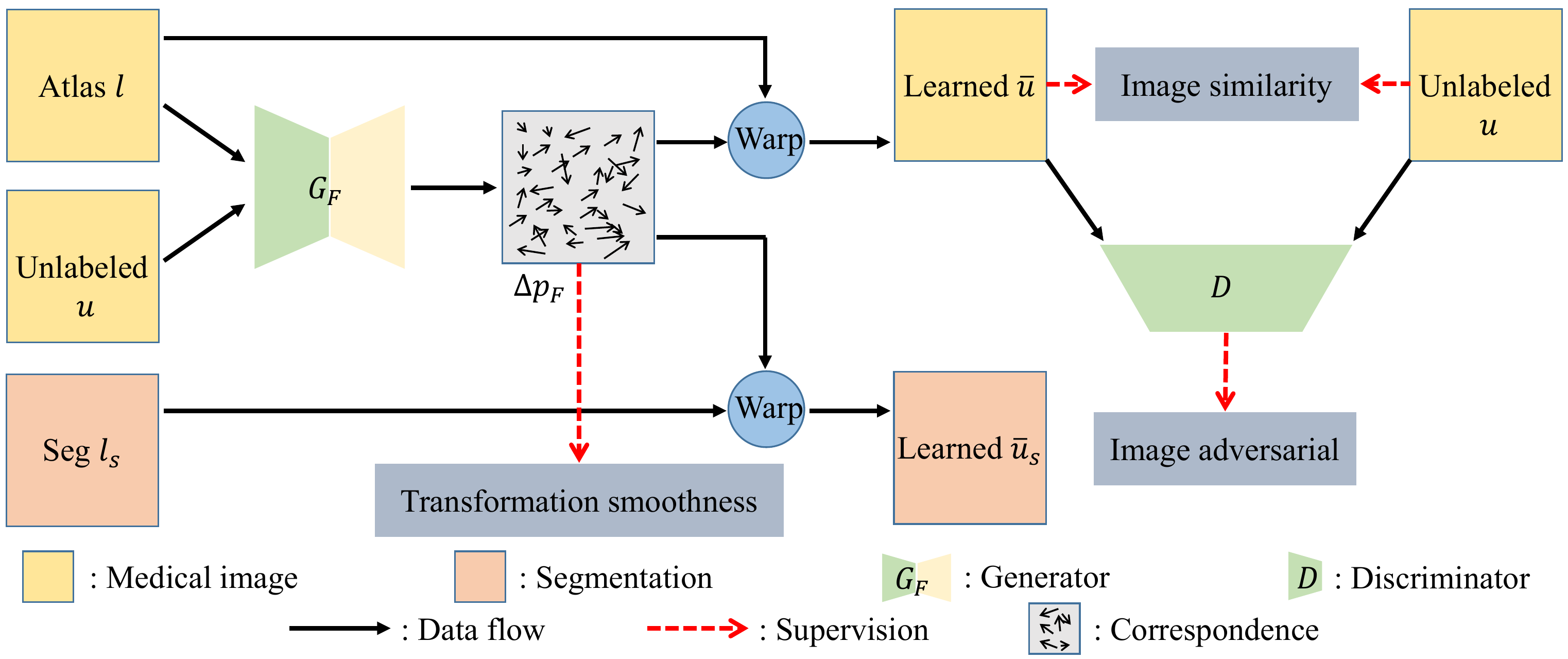}
\caption{The overview of the basic framework for correspondence learning. It consists of a generator network $G_F$ to learn the correspondences from the atlas to unlabeled images and two unsupervised losses, i.e., the image similarity loss $\mathcal{L}_\mathrm{sim}(\mathit{u},\bar{\mathit{u}})$ and the transformation smoothness loss $\mathcal{L}_\mathrm{smooth}(\Delta p_F)$. In addition, we introduce a generative adversarial network (GAN) \cite{goodfellow2014generative} into our basic framework to offer additional supervision.}
\label{fig_basic_framework}
\end{figure*}

This supplementary document provides more details about the basic framework for correspondence learning, in addition to the concise description in Section 3.

\textbf{The image similarity and transformation smoothness losses:}
As shown in Fig. \ref{fig_basic_framework}, to implement atlas-based segmentation with deep convolutional neural networks (\mbox{DCNNs}), a generator network $G_F$ is employed to learn the correspondences from the atlas to unlabeled images, and two unsupervised loss functions---the image similarity loss $\mathcal{L}_\mathrm{sim}(\mathit{u},\bar{\mathit{u}})$ and the transformation smoothness loss $\mathcal{L}_\mathrm{smooth}(\Delta p_F)$---are used to supervise the learning process.
Minimizing $\mathcal{L}_\mathrm{sim}$ encourages $\bar{\mathit{u}}$ to approximate
$\mathit{u}$, whereas minimizing $\mathcal{L}_\mathrm{smooth}$ regularizes $\Delta p_F$ to
be smooth.

To introduce robustness against global intensity variations in medical images caused by the differences in manufacturers, scanning protocols, and reconstruction methods, we adopt a locally normalized cross-correlation loss \cite{xu2019deepatlas,zhao2019dataaug} to formate $\mathcal{L}_\mathrm{sim}$ that encourages local coherence, which has been proven to be highly effective in correspondence-related tasks \cite{xu2019deepatlas,zhao2019dataaug}.
Let $f_u(t)$ and $f_{\bar{\mathit{u}}}(t)$ denote the functions to calculate local mean intensities of the unlabeled volume $\mathit{u}$ and deformed atlas $\bar{\mathit{u}}$: $f_u(t)=\frac{1}{n^3}\sum_{t_i}\mathit{u}(t_i)$ and $f_{\bar{\mathit{u}}}(t)=\frac{1}{n^3}\sum_{t_i}\bar{\mathit{u}}(t_i)$, where $t_i$ iterates over a $n^3$ cube around position $t$ in the volume, with $n=9$ in our experiments (the same as \cite{balakrishnan2019voxelmorph}). Then $\mathcal{L}_\mathrm{sim}(\mathit{u},\bar{\mathit{u}})$ is defined as:
\begin{equation}\label{eq:similarity_loss}
\begin{split}
&\mathcal{L}_\mathrm{sim}(\mathit{u}, \bar{\mathit{u}}) =  \\
&-\sum_{t\in\Omega}\frac{\Big(\sum_{t_i}\big(\mathit{u}(t_i)-f_u(t)\big)\big(\bar{\mathit{u}}(t_i)-f_{\bar{\mathit{u}}}(t)\big)\Big)^2}
{\Big(\sum_{t_i}\big(\mathit{u}(t_i)-f_u(t)\big)^2\Big)\Big(\sum_{t_i}\big(\bar{\mathit{u}}(t_i)-f_{\bar{\mathit{u}}}(t)\big)^2\Big)}.
\end{split}
\end{equation}

The smoothness constraint plays a key role in atlas-based segmentation methods \cite{xu2019deepatlas,zhao2019dataaug};
it is also widely used in other correspondence learning problems, such as optical flow estimation \cite{meister2018unflow, wang2018occlusion} and stereo matching \cite{lai2019bridging}. In addition, smoothness regularization can be considered as a strategy to alleviate the overfitting problem while encoding the anatomical priori.
Here, $\mathcal{L}_\mathrm{smooth}$ is formulated with the first-order derivative of $\Delta p_F$:
\begin{equation} \label{eq:smooth_loss}
\mathcal{L}_\mathrm{smooth}(\Delta p_F)= \sum_{ \mathit{t} \in \Omega}\Vert \nabla{(\Delta{p_F(t)})} \Vert_{2},
\end{equation}
where $t\in\Omega$ iterates over all spatial locations in $\Delta p_F$, and we approximate $\Vert \nabla{(\Delta{\mathit{p}(t)})} \Vert_{2}$ with spatial gradient differences between neighboring voxels along $x,y,z$ directions \cite{balakrishnan2019voxelmorph}:
\begin{equation} \label{eq:differential}
\begin{split}
\Vert \nabla{(\Delta{\mathit{p}(t)})} \Vert_{2} = &\frac{1}{3}(\Vert \nabla_x{(\Delta{\mathit{p}(t)})} \Vert_{2} +\\
&\Vert \nabla_y{(\Delta{\mathit{p}(t)})} \Vert_{2}+\Vert \nabla_z{(\Delta{\mathit{p}(t)})} \Vert_{2}).
\end{split}
\end{equation}

\textbf{The generative adversarial network (GAN) subnet:}
Besides $\mathcal{L}_\mathrm{sim}(\mathit{u},\bar{\mathit{u}})$ and $\mathcal{L}_\mathrm{smooth}(\Delta p_F)$---which are pretty much the standard configuration in atlas-based segmentation problems \cite{iglesias2015multi} (e.g., they were used as the main losses in VoxelMorph \cite{balakrishnan2019voxelmorph}), we introduce a GAN \cite{goodfellow2014generative} into our basic framework to offer additional supervision. The GAN subnet in our framework comprises $G_F$ and an additional discriminator network $D$ (see Fig. \ref{fig_basic_framework}).
A vanilla GAN would make the discriminator $D$ differentiate $\Delta{\mathit{p}_F}$ from the true underlying correspondence map. In practice, however, it is usually infeasible to obtain the true correspondence between a pair of clinical images.
Instead, we make $D$ distinguish $\bar{\mathit{u}}$ from $\mathit{u}$.
In this sense, $\bar{\mathit{u}}$ serves as a delegate of $\Delta{\mathit{p}_F}$, and $G_F$ is trained to generate $\Delta{\mathit{p}_F}$ that can be used to synthesize $\bar{\mathit{u}}$ authentically enough to confuse $D$;
meanwhile, $D$ becomes more skilled at flagging synthesized images.
This delegation strategy provides indirect supervision to $G_F$ and $\Delta p_F$, and allows the networks to be trained end-to-end with a large number of unlabelled images.

\begin{table}
	\caption{List of brain anatomical structures for segmentation from the CANDI dataset \cite{kennedy2011candishare}. `*/*' represents labels and categories which consist of left (L) and right (R) structures.
Abbreviations: white matter (WM), cortex (CX), ventricle (Vent), and cerebrospinal fluid (CSF).}\label{Category}
	\begin{center}\scalebox{0.93}{
		\begin{tabular}{c|c|c|c}
			Label & Category &Label &Category \\
			\hline
			2/41 & L/R-Cerebral-WM & 11/50 & L/R-Caudate\\
			3/42 & L/R-Cerebral-CX & 12/51 & L/R-Putamen\\
			4/43 & L/R-Lateral-Vent & 13/52 & L/R-Pallidum\\
			7/46 & L/R-Cerebellum-WM & 14 & $3^{rd}$-Vent\\
			8/47 & L/R-Cerebellum-CX & 15 & $4^{rd}$-Vent\\
		    10/49 & L/R-Thalamus-Proper & 16 & Brain-Stem\\
		    17/53 & L/R-Hippocampus & 24 & CSF\\
		    18/54 & L/R-Amygdala & 28/60 & L/R-VentralDC\\
		\end{tabular}}
	\end{center}
\end{table}

{\small
\bibliographystyle{ieee_fullname}
\bibliography{egbib}
}

\end{document}